%% file: main.tex
\documentclass{article}

\usepackage{microtype}
\usepackage{graphicx}
\usepackage{subcaption}
\usepackage{booktabs}
\usepackage{arydshln}
\usepackage{hyperref}
\usepackage[accepted]{icml2026}
\usepackage{amsmath}
\usepackage{amssymb}
\usepackage{mathtools}
\usepackage{amsthm}
\usepackage[capitalize,noabbrev]{cleveref}
\usepackage{multirow}
\usepackage{xcolor}
\usepackage{paralist}
\usepackage{enumitem}

\newcommand{\RTwo}{R^{2}}
\newcommand{\Af}{A_{f}}

\icmltitlerunning{Biokinetic Knowledge Priors for Data-Scarce Bioprocess Modeling}

\begin{document}

\twocolumn[
  \icmltitle{Leveraging Biokinetic Knowledge Priors\\for Data-Scarce Bioprocess Modeling}

  \icmlsetsymbol{equal}{*}

  \begin{icmlauthorlist}
    \icmlauthor{Kyunghoon Hur}{keti}
    \icmlauthor{Eunjung Jeon}{kfri}
    \icmlauthor{Hyun Woo Kim}{keti}
    \icmlauthor{Gyubok Lee}{kaist}
    \icmlauthor{Seongjun Yang}{cshl}
  \end{icmlauthorlist}

  \icmlaffiliation{keti}{Korea Electronics Technology Institute (KETI), Republic of Korea}
  \icmlaffiliation{kfri}{Korea Food Research Institute (KFRI), Republic of Korea}
  \icmlaffiliation{kaist}{Korea Advanced Institute of Science and Technology (KAIST), Republic of Korea}
  \icmlaffiliation{cshl}{Cold Spring Harbor Laboratory (CSHL), USA}

  \icmlcorrespondingauthor{Kyunghoon Hur}{hkhxoxo@gmail.com}

  \icmlkeywords{Bioprocess, Biokinetic ODE, Simulation Pre-training, AI for Science}

  \vskip 0.3in
]

\printAffiliationsAndNotice{}

\begin{abstract}
While deep learning has accelerated drug discovery, its impact on biomanufacturing has been considerably more limited.
The reason is data scarcity.
Bioreactor experiments are high-cost, take days to weeks, and are rarely shared in public form, leaving each research work with only a handful of experiments.
The domain itself, however, is rich in prior knowledge.
Biokinetic ordinary differential equation (ODE) models have described microbial growth for decades, yet how to inject this knowledge into a neural network has not been studied systematically.

We present the first systematic study of how to inject this ODE knowledge into a neural network, comparing a \emph{data-level prior} that pre-trains a generic decoder on simulated ODE curves against an \emph{architecture-level prior} that embeds the ODE inside the decoder.
Both consistently outperform no-prior baselines across $11$ datasets and $7$ microbial species.
Our central finding is that the two are \emph{substitutable}.
A generic decoder pre-trained on simulation matches a fully bio-structured decoder trained on real data.
Simulation pre-training therefore offers a simple, data-efficient recipe for deep learning under bioprocess data scarcity.
\end{abstract}

\section{Introduction}

\label{sec:intro}
\begin{figure*}[ht]
  \centering
  \includegraphics[width=0.90\textwidth]{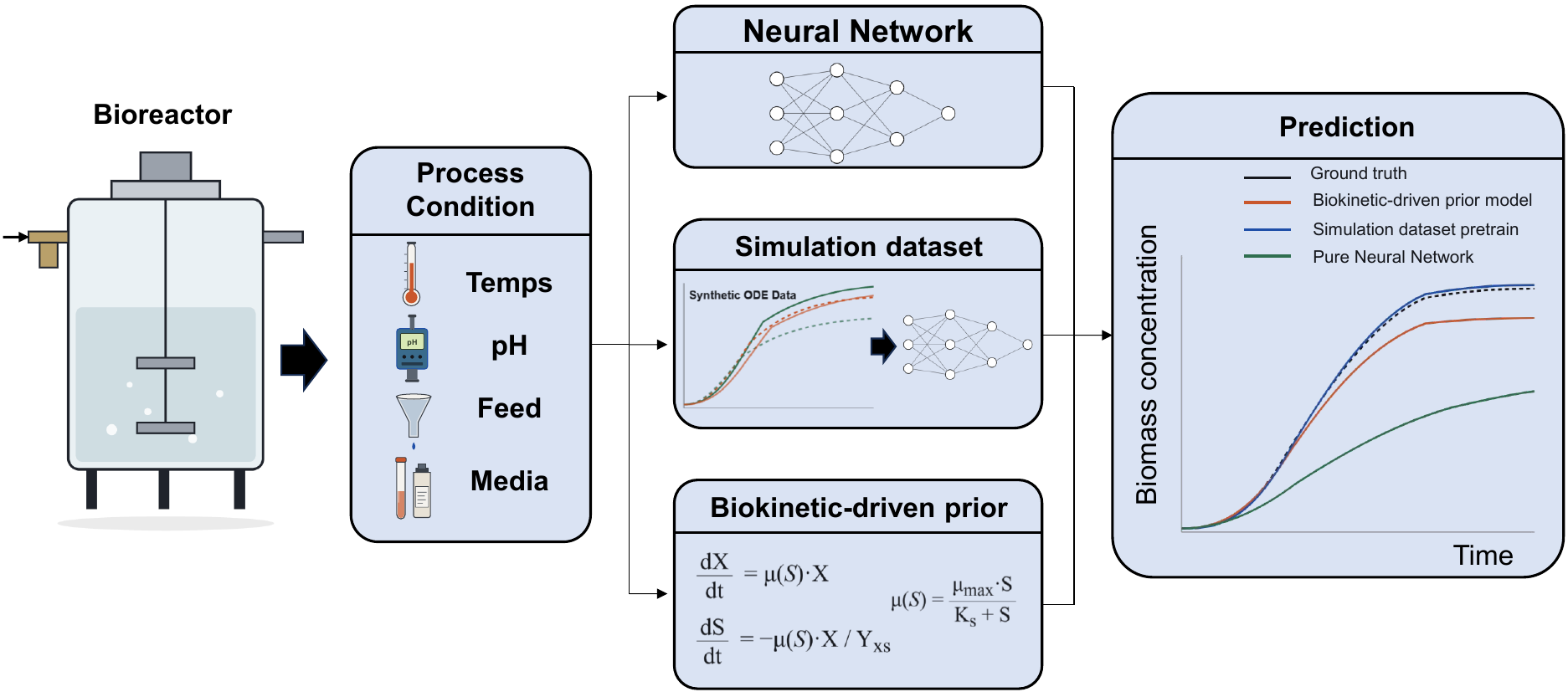}
  \caption{Task overview and the two biokinetic prior-injection channels compared in this paper. \textbf{(a) Task:} given environmental conditions $\mathbf{e}$ (e.g., temperature, pH, substrate, medium) and an optional set of early observations $\mathcal{C}$, predict the bioreactor state trajectory $\hat{y}(t)$ (cell density or an equivalent measurement). \textbf{(b) Two injection channels:} \textit{simulation pre-training} generates synthetic curves from parameterized biokinetic ODEs and uses them to pre-train a generic decoder, while an \textit{architecture-level prior} embeds the ODE directly in the decoder's forward pass. Section~\ref{sec:method} formalizes both channels on a shared encoder--decoder backbone.}
  \label{fig:overview}
\end{figure*}

\paragraph{Bottleneck in bioprocess AI}
Deep learning has driven major advances across the early stages of drug discovery~\citep{vamathevan2019applications, jimenez2020drug}.
Recent systems for biomolecular structure prediction~\citep{jumper2021highly, abramson2024accurate, lin2023evolutionary}, de novo protein design~\citep{watson2023novo}, antibiotic discovery~\citep{stokes2020deep}, and cellular simulation~\citep{bunne2024virtual} now contribute meaningfully to molecular design, target identification, and lead optimization.

The downstream stage of biomanufacturing, in which candidate molecules must be produced in bioreactors at scale, has by contrast attracted far less ML attention~\citep{mowbray2021,helleckes2023}.
A bioreactor experiment typically takes days to weeks and consumes substantial reagents, and every new product or strain demands re-tuning of cultivation conditions through trial and error~\citep{farid2020}.
Predicting bioreactor states such as cell density, substrate concentration, and product titer before an experiment is run would directly reduce cycle time and sample consumption, and would establish a substantive role for machine learning at this stage.

\paragraph{Research gap}
Three structural factors keep bioprocess modeling under-investigated by the ML community.
\textit{Data scarcity by experimental nature:} each cultivation is costly in both wall-clock time and reagents, so the data any single group can accumulate is intrinsically limited~\citep{barondiaz2025}.
\textit{Industrial confidentiality:} process data are treated as trade secrets, and very few datasets ever reach a public, curated form~\citep{smiatek2024}.
\textit{Method-level fragmentation:} as a consequence, datasets across groups serve different objectives, and direct method-to-method comparison is rarely performed; results from one group seldom transfer to data from other groups~\citep{khanal2024}.
Together, these factors explain why machine learning for bioprocess prediction has progressed unevenly to date.
Promising single-dataset case studies exist, but unified empirical baselines are rare.

\paragraph{Biokinetic ODE model}
Microbial growth dynamics, in contrast, have been described by biokinetic ODE models for several decades~\citep{monod1949growth, zwietering1990, baranyi1994dynamic, rosso1995}.
These models describe cell-level dynamics in closed form.
Once organism-specific parameters are fitted from the literature, simulation curves can be synthesized for any organism.
This makes biokinetic ODEs a rare resource in a data-scarce setting: prior knowledge that is mathematically precise, organism-portable, and freely available.
What remains open is a single design question: \emph{how should this knowledge be injected into a neural network?}

\paragraph{Two prior-injection channels}
We consider two orthogonal channels for injecting biokinetic ODE knowledge into a neural network, following the broader taxonomy of bias injection in physics-informed machine learning~\citep{karniadakis2021physics}.
The first is \emph{simulation pre-training} (a data-level prior), in which we generate large synthetic datasets from biokinetic ODEs, pre-train a decoder, and fine-tune on real data; pre-training on synthetic data has been a successful strategy in other data-scarce regimes~\citep{hollmann2023tabpfn}.
The second is an \emph{architecture-level prior}, in which the ODE itself is embedded in the model's forward pass as a template, hybrid, or Neural ODE backbone.

Prior work typically commits to one channel or the other.
Whether the two routes are complementary, substitutable, or one is strictly stronger has not been studied, even though they target the same underlying knowledge.
This paper is, to our knowledge, the first empirical study to compare the two channels under a single task, dataset suite, and shared backbone.
We also answer the practical question of \emph{how to construct an effective simulation dataset} (e.g., which ODE family, which parameter distribution, and how much data).
Figure~\ref{fig:overview} summarizes the task and the family of models we compare.

Our contributions are summarized as follows.
\begin{enumerate}[leftmargin=1.4em,itemsep=1pt,topsep=1pt]
  \item To our knowledge, this is the first empirical study that compares two channels for injecting biokinetic domain knowledge into neural networks, namely \emph{simulation pre-training} and \emph{architecture-level priors}, under a single task, shared backbone, and unified evaluation, across $11$ datasets and $7$ microbial species.
  \item Across these datasets, biokinetic priors consistently improve performance: both channels improve over no-prior baselines, and the improvement scales monotonically with prior intensity.
  \item Simulation pre-training is the more effective channel: a generic neural decoder paired with it matches a fully bio-structured decoder, demonstrating that the two channels are substitutable and that simulation is the more data-efficient route.
  \item Finally, we offer a practical recipe for constructing simulation datasets, with three findings: random simulation fails, composite-biokinetic simulation with broad parameter sampling performs best, and pre-training is at least as effective as joint training.
\end{enumerate}

\section{Related Work}
\label{sec:rw}

\paragraph{Data-driven ML/DL for bioprocess prediction}
Classical ML, such as PLSR, SVR, Gaussian processes, and XGBoost~\citep{peng2025,khuat2025}, and feed-forward MLPs (e.g., for CHO mAb titer~\citep{richter2025}) have been applied steadily to bioprocess prediction.
Sequence models such as the LSTM of \citet{bonanni2023} for \textit{E.~coli} OD600 forecasting (the basis of our GRU baseline) and Deep Set or autoencoder representations~\citep{borisyak2023,baig2023} extend the same data-driven paradigm.
These works learn the input--output mapping as a black box; biokinetic domain knowledge is rarely integrated explicitly.

\paragraph{Biokinetic models}
Mechanistic ODEs for microbial growth date back several decades.
Monod's saturation kinetics~\citep{monod1949growth}, Gompertz/Zwietering~\citep{zwietering1990}, Baranyi \& Roberts lag dynamics~\citep{baranyi1994dynamic}, and Luedeking--Piret product formation~\citep{luedeking1959} together describe biomass, substrate, and product dynamics through coupled mass-balance ODEs, and scale up to large process simulators such as IndPenSim~\citep{goldrick2015,goldrick2019}.
Their dominant application is in process control validation~\citep{li2024,petsagkourakis2020} or soft-sensor benchmarking~\citep{peng2025}; the use of biokinetic ODEs as a learning prior for neural networks has received little systematic attention.

\paragraph{Biokinetic-informed deep learning}
Existing approaches divide along three lines.
\textit{PINN (loss-level)} adds the biokinetic ODE residual as an auxiliary loss~\citep{adebar2025,zhu2026,kusters2025}.
\textit{Hybrid (architecture-level)} replaces reaction kinetics with a neural network on top of mass-balance ODEs~\citep{pinto2022,ramos2024}, or estimates time-varying parameters with a neural net~\citep{shah2022,riezzo2025}.
\textit{Neural ODE/UDE (solver-level)} places an ODE solver inside the forward pass~\citep{bangi2022,chiu2024}.
All of these works apply a single integration route to a single process; the gap we address is a systematic comparison of the two core channels (data-level simulation pre-training and architecture-level integration) on a single task, dataset suite, and shared backbone.

\section{Method}
\label{sec:method}

\subsection{Bioreactor state prediction}
\label{sec:method-task}

Public bioprocess datasets are dominated by microbial growth experiments, and fed-batch records are very rarely shared in a curated form, so we restrict our task formulation to batch microbial cultivation throughout the paper.
We formalize bioreactor state prediction in this regime as a conditional trajectory regression task.
Each cultivation run $i$ is described by a static \emph{environmental condition} vector $\mathbf{e}_i \in \mathbb{R}^{d_e}$, encoding cultivation parameters such as temperature, pH, initial substrate concentration, and medium composition.
Optionally, a set of \emph{context observations} $\mathcal{C}_i = \{(t_k, y_{i,k})\}_{k=1}^{K_i}$ provides up to $K_i$ early measurements collected during the same cultivation; $\mathcal{C}_i$ may be empty.
The model predicts the bioreactor state $\hat{y}_i(t) \in \mathbb{R}$ at any query time $t$, where $y$ denotes cell density or an equivalent measurement such as OD600.
Supervision is provided by ground-truth observations $y_i(t_j)$ on the time grid $\mathcal{T}_i = \{t_1, \ldots, t_{T_i}\}$ at which the cultivation was sampled.

Every model in this paper factorizes the predictor into the same two-stage mapping,
\begin{equation}
\mathbf{z}_i \;=\; \mathrm{Encoder}\!\left(\mathbf{e}_i,\, \mathcal{C}_i\right),
\qquad
\hat{y}_i(t) \;=\; \mathrm{Decoder}\!\left(\mathbf{z}_i,\, t\right),
\label{eq:two-stage}
\end{equation}
where $\mathbf{z}_i \in \mathbb{R}^{d_z}$ is a per-cultivation latent representation produced by a shared encoder (EnvEncoder $+$ ContextEncoder; see Appendix~\ref{sec:appendix-arch}).
Only the decoder differs from one baseline to another, which isolates the architectural prior as the only source of variation across models.

Models are trained against the trajectory regression loss
\begin{equation}
\mathcal{L}_i \;=\; \frac{1}{T_i}\sum_{j=1}^{T_i} \ell\!\left(\hat{y}_i(t_j),\; y_i(t_j)\right),
\label{eq:traj-loss}
\end{equation}
where $\ell(\cdot,\cdot)$ is a per-point regression loss.
The simplest concrete instance is the squared error $\ell(\hat{y}, y) = (\hat{y} - y)^{2}$, which yields the mean-squared-error (MSE) loss.
All baselines compared in this paper share this trajectory regression form; the specific choice of $\ell$ and the evaluation metrics are defined in Section~\ref{sec:exp}.

\subsection{Biokinetic prior--injected models}
\label{sec:method-models}

\paragraph{Component-based hybrid family}
Every prior-injected baseline in this paper fits a unified template: a biokinetic prior plus a small neural correction,
\begin{equation}
\hat{y}_i(t) \;=\; f_{\mathrm{prior}}\!\big(t;\, \boldsymbol{\theta}_i\big) \;+\; \varepsilon \cdot g_\theta\!\big(\mathrm{state},\, t,\, \mathbf{z}_i\big),
\label{eq:hybrid-form}
\end{equation}
where $\boldsymbol{\theta}_i = \mathrm{ParamHead}(\mathbf{z}_i)$ and $\varepsilon \in \{0.01, 0.1\}$.
Here $f_{\mathrm{prior}}$ is a biokinetic template (a closed form or an ODE solution), $g_\theta$ is a small neural residual, and $\boldsymbol{\theta}_i$ collects the organism-specific dynamic parameters (e.g., $\mu_{\max}$, $K_s$, $K$, $k_d$) emitted by ParamHead from the latent $\mathbf{z}_i$.
The scalar $\varepsilon$ controls the influence of the neural correction: a small $\varepsilon$ favors the prior, while a large $\varepsilon$ favors the neural network.
Within this template we instantiate four representative architectures whose prior intensity increases from left to right.

\paragraph{\textit{MLP}}
We start at the no-prior end of the spectrum.
This baseline is the $f_{\mathrm{prior}} \equiv 0$ limit of \cref{eq:hybrid-form}, in which the decoder is a $4$-layer fully-connected ReLU network with no ODE structure:
\begin{equation}
\hat{y}_i(t) \;=\; \mathrm{MLP}\!\big(\mathbf{z}_i,\, t\big).
\label{eq:mlp}
\end{equation}
The MLP carries no biokinetic structure and acts as the no-prior reference point in our comparison.

\paragraph{\textit{PINN}}
At a higher level of prior intensity, we retain the same backbone and incorporate the biokinetic ODE into the loss instead of the architecture.
Building on the physics-informed neural network framework~\citep{raissi2019physics} and following its bioprocess instantiation in~\citet{adebar2025}, the PINN baseline uses the same MLP backbone as above and adds an autograd-based logistic-residual term to the loss:
\begin{align}
\mathcal{L}^{\mathrm{PINN}}_i \;=\;& (1-\lambda)\,\mathcal{L}_i^{\mathrm{MSE}} \;+\; \frac{\lambda}{T_i}\sum_{j=1}^{T_i} \nonumber \\
&\Big| \tfrac{d\hat{y}_i}{dt}(t_j) - \mu\,\hat{y}_i(t_j)\!\left(1 - \tfrac{\hat{y}_i(t_j)}{K}\right) \Big|^{2}.
\label{eq:pinn}
\end{align}
The derivative $d\hat{y}_i/dt$ is evaluated by autograd, the trade-off coefficient is fixed at $\lambda = 0.1$, and the dataset-level $(\mu, K)$ are taken from the same Gompertz fits used by ODE-Fit so that the loss-channel and the template-channel see the same kinetic prior.
The architecture is unchanged, but the model is now regularized toward biokinetic dynamics during training.

\paragraph{\textit{Hybrid-NeuralODE}}
Pushing the prior further into the architecture, we embed the ODE itself inside the forward pass and let a small neural network correct it~\citep{chen2018neural}.
Following~\citet{bangi2022}, this baseline integrates a Baranyi two-state system,
\begin{align}
\frac{dN_i}{dt} &= \mu \cdot \frac{q_i}{1+q_i} \cdot \Big(1 - \tfrac{N_i}{K}\Big) \cdot N_i + \varepsilon\, g_\theta(N_i,t,\mathbf{z}_i), \nonumber \\
\frac{dq_i}{dt} &= \mu \cdot q_i,
\label{eq:hybrid-neuralode}
\end{align}
where $N_i$ is cell density, $q_i$ is the Baranyi lag adaptation state, and $\mu$ is treated as a constant (substrate dynamics are not modeled here).
The model output is now an ODE solution, and the neural correction $g_\theta$ modulates that solution only marginally.

\paragraph{\textit{BioStruct-ODE family}}
At the strong-prior end of the spectrum, we introduce the BioStruct-ODE family, which adopts the Monod--Baranyi system as the model's backbone.
The system tracks four states explicitly, namely cell density $N_i$, substrate $S_i$, product $P_i$, and lag adaptation $q_i$, and includes an explicit death term:
\begin{align}
\frac{dN_i}{dt} &= \mu(S_i)\!\cdot\!\frac{q_i}{1+q_i}\!\cdot\!\Big(1 - \tfrac{N_i}{K}\Big) N_i - k_d N_i + \varepsilon\, g_\theta, \nonumber \\
\frac{dS_i}{dt} &= -\,\frac{\mu(S_i)}{Y_{xs}}\, N_i + \varepsilon\, g_\theta, \nonumber \\
\frac{dP_i}{dt} &= \alpha\,\mu(S_i)\, N_i + \beta\, N_i + \varepsilon\, g_\theta, \nonumber \\
\frac{dq_i}{dt} &= \mu(S_i)\cdot q_i.
\label{eq:biostruct}
\end{align}
The growth rate $\mu(S_i)$ follows Monod kinetics, and the bias of ParamHead is \emph{warm-started} from per-organism literature Monod fits to stabilize training (Appendix~\ref{sec:appendix-arch}).
The family contains two variants.
\textit{BioStruct-ODE} uses the system above as is.
\textit{BioStruct-ODE-Cardinal} additionally models the dependence of $\mu$ on temperature and pH through Rosso's cardinal envelope~\citep{rosso1995},
\begin{equation}
\mu(S_i, T_i, \mathrm{pH}_i) \;=\; \gamma(T_i)\cdot \gamma(\mathrm{pH}_i)\cdot \mu_{\max}\cdot \frac{S_i}{K_s + S_i},
\label{eq:cardinal}
\end{equation}
where $\gamma(\cdot)$ is the cardinal envelope defined by the cardinal points $(T_{\min},T_{\mathrm{opt}},T_{\max})$ and $(\mathrm{pH}_{\min},\mathrm{pH}_{\mathrm{opt}},\mathrm{pH}_{\max})$.
The Cardinal variant is applied only to datasets with an explicit environmental axis; datasets without such an axis fall back to BioStruct-ODE automatically.

\subsection{Training with simulated datasets}
\label{sec:method-sim}

\paragraph{Simulation data synthesis}
We sample organism-specific parameters from literature ranges and integrate the corresponding ODE family to obtain simulation curves.
Three design factors, ablated in RQ2, control the resulting synthetic dataset.
\textit{Factor A (ODE family)} chooses among Monod, Logistic, Gompertz, Baranyi, and a Rosso-composite form.
\textit{Factor B (parameter sampling)} contrasts a literature-narrow regime ($\pm$10\%) with a broad uniform regime ($\pm$50\%).
\textit{Factor C (biokinetic specificity)} arranges three regimes in increasing specificity: random Gaussian processes (no biokinetic structure), single-ODE narrow, and composite-ODE broad.
Full ODE equations and per-organism parameter ranges are deferred to Appendix~\ref{sec:appendix-sim}.

\paragraph{Pre-training}
We train on simulation to convergence and then fine-tune on real data with a smaller learning rate.
The simulation budget is scaled relative to the real dataset, so that when real data is scarce, more simulation is used to cover a broader biokinetic regime; we sweep the ratio $|\mathrm{sim}|/|\mathrm{real}| \in \{1, 3, 5, 10, 30\}$.

\paragraph{Joint training}
Simulation and real samples are mixed within each batch and weighted by a curriculum coefficient $\alpha(t)$ that decays from $1$ (simulation-dominated) to $0$ (real-only) as training progresses,
\begin{equation}
\mathcal{L}_{\mathrm{joint}} \;=\; \alpha(t)\,w_{\mathrm{sim}}\,\mathcal{L}_{\mathrm{sim}} + (1 - \alpha(t))\,\mathcal{L}_{\mathrm{real}},
\label{eq:joint}
\end{equation}
with $w_{\mathrm{sim}} \in \{0.5, 1, 2, 5\}$.
Optimiser, schedule, loss-term, and batch-composition details appear in Appendix~\ref{sec:appendix-training}.

\section{Experiments}
\label{sec:exp}

In this section we address four research questions.
\begin{itemize}[leftmargin=1.4em,itemsep=1pt,topsep=1pt]
  \item \textbf{RQ1.} How do different ways of injecting an architectural biokinetic prior compare in performance?
  \item \textbf{RQ2.} How should simulation data be synthesized, and how should it be injected into the model (pre-training vs.\ joint training)?
  \item \textbf{RQ3.} How does test-time context (initial conditions only vs.\ partial observations) affect performance?
  \item \textbf{RQ4.} What kinds of growth-curve errors does simulation pre-training systematically correct?
\end{itemize}

\paragraph{Datasets}
We curate $11$ datasets covering $7$ bacterial species (\textit{E.~coli}, \textit{Listeria}, \textit{Shigella}, \textit{Staphylococcus aureus}, \textit{Yersinia}, \textit{Salmonella}, and Pseudomonads), comprising two batch datasets of our own~\citep{ucl_rdr_bl21,katipogluyazan2023} and nine external datasets~\citep{faure2023,buchanan1989,zaika1994,eifert1997,combase}.
Per-dataset organism, curve count, time horizon, and split are reported in Appendix~\ref{sec:appendix-dataset}.

\paragraph{Pre-processing}
OD-native datasets (BL21, Katipoglu23, Faure23) are normalized in OD600 units, while ComBase data are kept in their native logCFU units.
We use a curve-level random train/val/test split, and apply a \texttt{with\_growth} filter that removes any curve whose per-curve Monod--Baranyi NLS fit attains $\RTwo < 0.4$.
Details are in Appendix~\ref{sec:appendix-dataset}.

\paragraph{Models and implementation}
All baselines share the same backbone (EnvEncoder $+$ ContextEncoder), and only the decoder changes.
Each model is trained with $5$ seeds.
The default per-point loss is shared across baselines, and PINN additionally uses the auxiliary physics-residual term in \cref{eq:pinn}; full loss forms are listed in Appendix~\ref{sec:appendix-loss}.
We optimize with AdamW (lr $3 \times 10^{-4}$, weight decay $10^{-5}$, $10$-epoch warmup), and use early stopping on validation $\RTwo$ with patience $5$.
Further detail is in Appendices~\ref{sec:appendix-arch} and~\ref{sec:appendix-training}.

\paragraph{Baselines}
We evaluate nine baselines.
\textbf{\textit{Mean}} predicts the per-condition mean trajectory of the training set.
\textbf{\textit{ODE-Fit}} performs per-curve nonlinear least-squares fits with Monod or Gompertz forms~\citep{zwietering1990}.
\textbf{\textit{MLP}} is a pure feed-forward decoder.
\textbf{\textit{GRU}} is an autoregressive recurrent decoder following~\citet{bonanni2023}.
\textbf{\textit{PINN}} pairs the MLP backbone with the auxiliary physics loss of~\citet{adebar2025}.
\textbf{\textit{ODE-Guide}} combines a closed-form Gompertz template with a neural residual~\citep{pinto2022}.
\textbf{\textit{Hybrid-NeuralODE}} combines a mass-balance ODE with a neural correction~\citep{bangi2022}.
The two main variants of this paper are \textbf{\textit{BioStruct-ODE}} and \textbf{\textit{BioStruct-ODE+Cardinal}} (Section~\ref{sec:method-models}); the Cardinal variant requires an explicit temperature axis in the dataset, so we evaluate it only on the $7$ datasets that carry one, while every other baseline is evaluated on all $11$.
Architectural details are in Appendix~\ref{sec:appendix-arch}.

\input{tables/t1_rq1}

\paragraph{Evaluation}
We report five metrics:
\begin{itemize}[leftmargin=1.4em,itemsep=1pt,topsep=1pt]
  \item $\RTwo$ (primary, dimensionless), the proportion of trajectory variance explained by the prediction;
  \item $\Af$ (Accuracy Factor, $\Af = 10^{\overline{|\log_{10}(\hat{y}/y)|}}$;~\citealp{ross1996}), the average fold-change between predicted and observed values;
  \item RMSE in log10 cell density, ensuring fair comparison across native units;
  \item EndPntErr in OD600 at the final timepoint, with eval-time per-organism offsets for ComBase;
  \item Rank, the average rank of each model across $(\text{dataset}, \text{seed})$ pairs.
\end{itemize}
$\RTwo$, $\Af$, and Rank are scale-invariant; RMSE is computed in log space; EndPntErr is computed in OD600, where endpoints cluster at the plateau OD~$\sim$0.3--1.0.

\subsection{RQ1: Comparison of prediction models with biokinetic prior}
\label{sec:rq1}

\textbf{A unified comparison of nine baselines}
Prior work studies the five architectural routes (loss-level, template, hybrid, Neural ODE, full bio-structured) in isolation, on different datasets, with different encoders and evaluation protocols, which makes their relative strength impossible to infer from the published numbers.
We therefore evaluate nine baselines on the full $11$-dataset suite for $5$ seeds with no simulation, sharing the unified backbone of Section~\ref{sec:method} so that the only varying factor across rows is the decoder; results are reported in Table~\ref{tab:t1-rq1}.

\textbf{Prior intensity orders performance}
The BioStruct-ODE family attains the best mean $\RTwo$ ($0.521$ for the base variant and $0.554$ for the Cardinal variant on its seven temperature-axis datasets) and the best average rank ($3.00$ and $2.76$ out of nine), outperforming every other baseline.
The loss-level prior (PINN, $\RTwo = 0.503$) and the sequence-bias channel (GRU, $\RTwo = 0.494$) cluster just below the BioStruct-ODE family.
Mean ($0.157$), out-of-sample ODE-Fit ($0.204$), and ODE-Guide ($0.090$) rank at the bottom: a biokinetic template without data-driven adaptation (ODE-Fit) only barely exceeds the trivial Mean baseline, while a closed-form prior with only a weak neural correction (ODE-Guide) performs worse than the Mean baseline.

\textbf{Variance tightens, and component ablation supports it}
The no-prior MLP exhibits a per-curve standard deviation of $1.20$ that is dominated by a small number of difficult datasets (notably Zaika~$1994$, where the unconstrained network produces strongly negative $\RTwo$), while the BioStruct-ODE family suppresses this to about $0.21$.
The biokinetic component ablation in Appendix~\ref{sec:appendix-biokinetic-ablation} (Table~\ref{tab:appendix-biokinetic-component}) supports this attribution: removing the death term, the warm-start, or switching the template from Monod--Baranyi to Gompertz each incurs a small but consistent loss in both $\RTwo$ and $\Af$.
Architecture-level injection therefore helps under data scarcity, and this reframes the central question of the paper: if architecture-level injection improves performance, can the same knowledge be supplied through simulation instead, and how do the two channels relate?

\begin{figure*}[ht]
\centering
\includegraphics[width=0.95\textwidth]{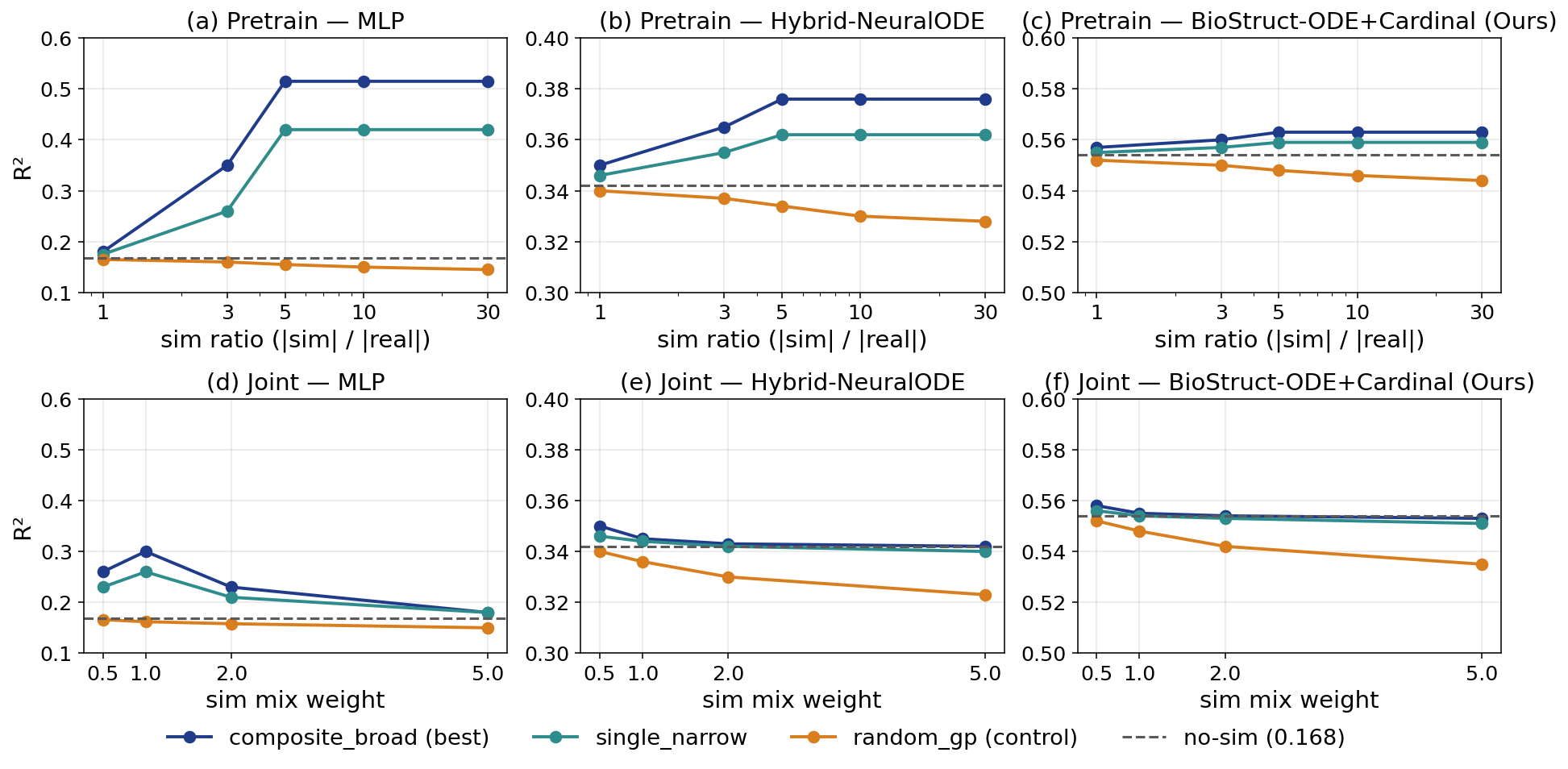}
\caption{Effect of simulation methodology and injection mode on test $\RTwo$, across three architectures spanning the prior-intensity spectrum (MLP, Hybrid-NeuralODE, BioStruct-ODE+Cardinal). The top row sweeps pre-training sim ratio $|\mathrm{sim}|/|\mathrm{real}| \in \{1, 3, 5, 10, 30\}$, and the bottom row sweeps joint-training sim mix weight $\in \{0.5, 1, 2, 5\}$. Each line depicts a sim variant (composite-biokinetic broad, single-ODE narrow, random-GP control); dashed horizontal lines depict the corresponding no-sim baselines that match those in Table~\ref{tab:t1-rq1}. Values are averaged over 5 seeds and the $y$-axis is per-column.}
\label{fig:f3-sim-ablation}
\end{figure*}

\subsection{RQ2: Simulation prior injection}
\label{sec:rq2}

\textbf{Simulation as a second injection route}
The same ODEs that we embed in the decoder for BioStruct-ODE can also be sampled and integrated to produce a synthetic dataset of curves, which then pre-trains or jointly trains an otherwise generic decoder.
We therefore ask two coupled questions: \emph{how} should the synthetic dataset be constructed, and \emph{how} should it be injected into training?
We select three architectures spanning the prior-intensity spectrum (MLP, Hybrid-NeuralODE, BioStruct-ODE+Cardinal) and compare two injection modes: (i) pre-training at five sim ratios and (ii) joint training at four mix weights.
Each setting is evaluated with three sim variants: composite-biokinetic broad sampling, single-ODE narrow sampling, and a non-biokinetic random-GP control that removes biokinetic structure while preserving curve smoothness.
Results are reported in Figure~\ref{fig:f3-sim-ablation}.

\textbf{Biokinetic specificity is the key factor}
Inspecting Figure~\ref{fig:f3-sim-ablation} column by column, every architecture exhibits the same ordering: composite-biokinetic broad sampling outperforms single-ODE narrow sampling, and both outperform the random-GP control by a margin that widens with the simulation budget.
Smoothness and curve-shape regularity therefore cannot explain the gain; the simulation must carry biokinetic structure (saturation, lag, carrying capacity) for the prior to transfer.

\textbf{Pre-training improves MLP the most}
The MLP column of Figure~\ref{fig:f3-sim-ablation} exhibits the largest improvement: composite-broad pre-training improves $\RTwo$ from the no-sim baseline of $0.168$ to $\approx 0.515$ at saturation, on par with a BioStruct-ODE+Cardinal trained from scratch on real data alone ($0.554$ on its seven-dataset subset).
Joint training, by contrast, peaks at low mix weights and then degrades; for an MLP the joint maximum is $\RTwo \approx 0.30$ at mix weight $1$ before declining toward the no-sim baseline at higher weights, whereas pre-training saturates and remains stable.

\textbf{A practical recipe for the data-scarce setting}
The match between MLP$+$pre-train ($\RTwo \approx 0.515$) and BioStruct-ODE+Cardinal$+$nosim ($0.554$) is our key substitution finding: the architecture and data channels behave as substitutes for the same biokinetic prior, and the gap between them lies within the seed-level standard deviations reported in Table~\ref{tab:t1-rq1}.
We therefore recommend composite-biokinetic broad pre-training at moderate sim ratios as a practical, data-efficient strategy for practitioners with a generic decoder and limited real data.

\begin{figure}[ht]
\centering
\includegraphics[width=0.90\columnwidth]{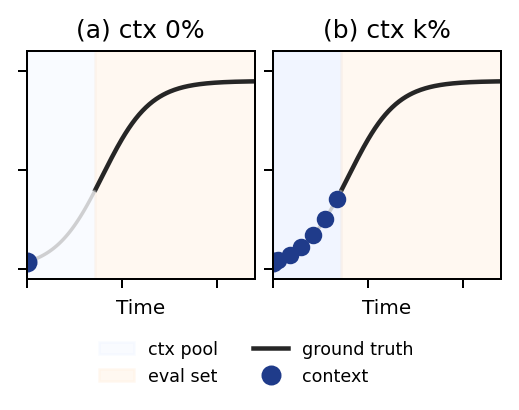}
\caption{Illustration of the test-time context-conditioning protocol used in RQ3 (schematic, not measured). A fixed window of timepoints (orange band) is held out as the evaluation set across all context densities, while context observations (filled circles) are drawn only from the complementary context pool (blue band, depicted only when context is supplied).}
\label{fig:f4-ctx-protocol}
\end{figure}

\subsection{RQ3: Context conditioning}
\label{sec:rq3}

\textbf{Does context replace priors?}
To make different context densities directly comparable, we split each test curve into a fixed back-$70\%$ evaluation window and a front-$30\%$ context pool.
Every baseline is then evaluated at three context densities ($0$, $10$, and $30\%$ of the curve), with the back-$70\%$ window held fixed across all densities so that only the inputs received during inference change.
The protocol is depicted in Figure~\ref{fig:f4-ctx-protocol} and the results are reported in Table~\ref{tab:t2-rq3}.

\textbf{Context helps; pre-training transfers cleanly}
Across all eight trainable baselines in Table~\ref{tab:t2-rq3}, moving from ctx-$0\%$ to ctx-$10\%$ yields a small but consistent gain (mean $\Delta\RTwo \approx +0.012$, mean $\Delta\Af \approx -0.010$), and the additional gain from ctx-$10\%$ to ctx-$30\%$ is smaller still ($\Delta\RTwo \approx +0.005$).
Every $+$Pretrain sub-row outperforms its scratch counterpart at every context density, with the largest absolute gain on the weakest backbone (MLP$+$Pretrain at ctx-$0\%$ improves $\RTwo$ by about $+0.347$ over MLP scratch) and a much smaller gain on the strongest one (BioStruct-ODE+Cardinal$+$Pretrain at ctx-$0\%$ improves it by about $+0.009$).

\textbf{Architecture, simulation, and context stack additively}
The strongest single configuration is BioStruct-ODE+Cardinal$+$Pretrain at ctx-$30\%$, attaining $\RTwo = 0.577$, which improves on the strongest scratch configuration at the same context density ($0.564$) and on the no-context Cardinal+Pretrain ($0.563$).
In practice, $\approx 10\%$ of the curve is the operating point with the best gain-per-cost trade-off, while simulation pre-training and an architectural prior should still be applied on top of context.
\input{tables/t2_rq3}

\subsection{RQ4: Failure-mode analysis}
\label{sec:rq4}

\begin{figure}[ht]
\centering
\includegraphics[width=0.98\columnwidth]{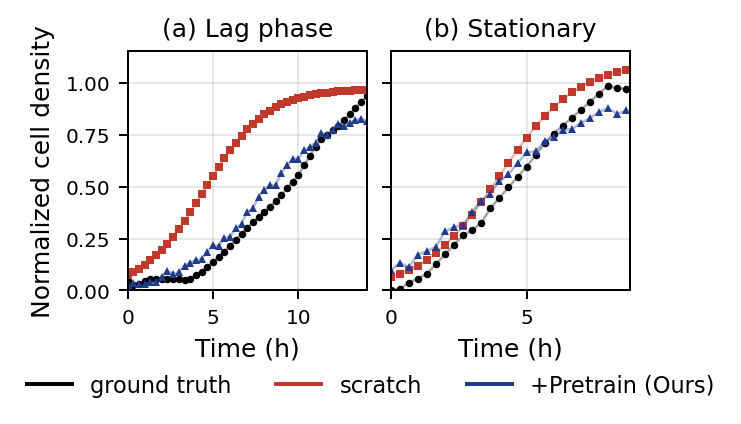}
\caption{Failure-mode comparison of the scratch and $+$Pretrain predictions of BioStruct-ODE+Cardinal on two hand-selected real test curves. \textbf{(a)} Lag phase, drawn from \textit{Listeria}~Buchanan~$1989$ at $5\,^{\circ}$C: the scratch model predicts immediate growth, while $+$Pretrain captures the lag. \textbf{(b)} Stationary phase, drawn from ComBase~\textit{Yersinia} at $0\,^{\circ}$C: the scratch model overshoots the plateau, while $+$Pretrain converges near the correct level (Section~\ref{sec:rq4}).}
\label{fig:f5-failure-modes}
\end{figure}

\textbf{What kind of error does pre-training correct?}
To investigate this further, we evaluate the strongest configuration (BioStruct-ODE-Cardinal$+$Pretrain) on two real test curves and contrast its predictions with the same model trained from scratch (Figure~\ref{fig:f5-failure-modes}).

\textbf{Lag and carrying capacity recovered}
On a low-temperature \textit{Listeria} lag-phase example (Figure~\ref{fig:f5-failure-modes}a), the scratch model predicts immediate growth from time zero and overshoots the early measurements with an almost monotonic curve.
$+$Pretrain, by contrast, remains near zero growth through the first few hours and only then begins to increase, tracking the ground truth.
On a \textit{Yersinia} stationary-phase example (Figure~\ref{fig:f5-failure-modes}b), the scratch model overshoots the plateau and continues to climb above the carrying capacity, while $+$Pretrain approaches the correct $K$ and remains at the plateau.

\textbf{Errors map onto specific ODE terms}
The lag correction corresponds to the lag dynamics $q$ in the BioStruct-ODE system, and the plateau correction corresponds to the carrying capacity $K$; a similar argument extends in principle to the death term $k_d$ in the late phase of the curve.
A generic smoothing prior would not isolate these particular ODE terms but would instead shrink the prediction toward an arbitrary mean trajectory.

\section{Conclusion}
\label{sec:conclusion}

Biokinetic-ODE-derived simulation curves can substitute for a substantial portion of scarce real bioprocess data.
A unified comparison demonstrates that simulation pre-training and architecture-level priors act as substitutable channels, with biokinetic specificity as the key factor.
In particular, a generic neural decoder with simulation pre-training matches a fully bio-structured decoder, while random-curve simulation fails and pre-training outperforms joint training.

\paragraph{Limitations}
Cross-organism transfer to unseen species remains unsolved (Appendix~\ref{sec:appendix-cross-organism}), and our scope excludes mammalian cell systems where industrial data privacy dominates.
Our evaluation is limited to bacterial batch cultivation; fed-batch dynamics, perfusion, and continuous culture are not addressed.
We use $5$ seeds per configuration, which limits the precision of variance estimates for borderline comparisons.
The RQ4 mechanism evidence is illustrative rather than aggregate, and the average-case quantitative evidence is concentrated in Tables~\ref{tab:t1-rq1}--\ref{tab:t2-rq3} and Figure~\ref{fig:f3-sim-ablation}.

\paragraph{Future work}
A natural next step is to extend the prediction interface into closed-loop deployment with dynamic process control or online monitoring systems.
The substitutability finding also suggests hybrid pipelines that allocate effort between architecture engineering and simulation generation according to the available compute and engineering cost.

\section{Dataset availability}
\label{sec:availability}

All datasets used in this work are publicly available.

\section*{Institutional Review Board (IRB)}

This research does not require IRB approval.

\section*{Acknowledgments}

This work was supported by Institute of Information \& communications Technology Planning \& Evaluation (IITP) grant funded by the Korea government(MSIT) (RS-2024-00346798,

\clearpage
\bibliographystyle{icml2026}
\bibliography{refs}

\clearpage
\appendix
\input{appendix}

\end{document}

%% file: tables/t1_rq1.tex
\begin{table*}[!ht]
\centering
\caption{Per-model performance on the $11$-dataset suite at zero context (RQ1). Each model is trained for $5$ seeds without simulation pre-training; values are mean~$\pm$~standard deviation, and the Rank column reports each model's rank averaged across datasets. The best value in every column is shown in bold.}
\label{tab:t1-rq1}
\begin{tabular}{lccccc}
\toprule
Model & $\RTwo \uparrow$ & $\Af \downarrow$ & RMSE$_{\log} \downarrow$ & EndPntErr$_{\mathrm{OD}} \downarrow$ & Rank $\downarrow$ \\
\midrule
Mean              & $0.157 \pm 0.336$ & $1.418 \pm 0.253$ & $1.597 \pm 0.927$ & $0.377 \pm 0.232$ & $6.82$ \\
ODE-Fit           & $0.204 \pm 0.342$ & $1.338 \pm 0.136$ & $1.526 \pm 0.891$ & $0.491 \pm 0.267$ & $6.79$ \\
\midrule
MLP               & $0.168 \pm 1.203$ & $1.329 \pm 0.149$ & $1.420 \pm 0.855$ & $0.447 \pm 0.282$ & $4.82$ \\
GRU               & $0.494 \pm 0.246$ & $1.288 \pm 0.153$ & $1.197 \pm 0.692$ & $0.401 \pm 0.248$ & $3.55$ \\
\midrule
PINN              & $0.503 \pm 0.188$ & $1.287 \pm 0.135$ & $1.196 \pm 0.674$ & $0.413 \pm 0.252$ & $3.45$ \\
ODE-Guide         & $0.090 \pm 0.346$ & $1.437 \pm 0.205$ & $1.507 \pm 0.866$ & $0.422 \pm 0.231$ & $6.61$ \\
Hybrid-NeuralODE  & $0.342 \pm 0.349$ & $1.336 \pm 0.139$ & $1.388 \pm 0.833$ & $0.399 \pm 0.239$ & $4.94$ \\
\midrule
BioStruct-ODE (Ours)        & $0.521 \pm 0.254$ & $1.266 \pm 0.136$ & $\mathbf{1.167 \pm 0.695}$ & $\mathbf{0.372 \pm 0.235}$ & $3.00$ \\
\hspace{1em}$+$Cardinal     & $\mathbf{0.554 \pm 0.212}$ & $\mathbf{1.235 \pm 0.079}$ & $1.322 \pm 0.577$ & $0.432 \pm 0.237$ & $\mathbf{2.76}$ \\
\bottomrule
\end{tabular}
\end{table*}

%% file: tables/t2_rq3.tex
\begin{table*}[!ht]
\centering
\caption{Test-time $\RTwo$ and $\Af$ at three context densities (RQ3). The \textit{$+$Pretrain} sub-rows show the best simulation configuration (composite-biokinetic broad pre-train, sim\_ratio $=10$) for the three representative architectures. The ctx-$0\%$ column for scratch rows matches Table~\ref{tab:t1-rq1} exactly. Values are mean${}_{\pm\text{std}}$ across $5$ seeds, and the best value in each column is shown in bold.}
\label{tab:t2-rq3}
\fontsize{9}{11}\selectfont
\setlength{\dashlinedash}{0.6pt}
\setlength{\dashlinegap}{1.5pt}
\begin{tabular}{l ccc :ccc}
\toprule
& \multicolumn{3}{c:}{$\RTwo \uparrow$} & \multicolumn{3}{c}{$\Af \downarrow$} \\
\cmidrule(lr){2-4} \cmidrule(lr){5-7}
Model & $0\%$ & $10\%$ & $30\%$ & $0\%$ & $10\%$ & $30\%$ \\
\midrule
Mean             & $0.157_{\pm .336}$ & $0.157_{\pm .336}$ & $0.157_{\pm .336}$ & $1.418_{\pm .253}$ & $1.418_{\pm .253}$ & $1.418_{\pm .253}$ \\
ODE-Fit          & $0.204_{\pm .342}$ & $0.215_{\pm .330}$ & $0.220_{\pm .325}$ & $1.338_{\pm .136}$ & $1.328_{\pm .134}$ & $1.322_{\pm .132}$ \\
\midrule
MLP              & $0.168_{\pm 1.20}$ & $0.180_{\pm 1.10}$ & $0.185_{\pm 1.05}$ & $1.329_{\pm .149}$ & $1.318_{\pm .145}$ & $1.312_{\pm .142}$ \\
\hspace{1em}$+$Pretrain & $0.515_{\pm .205}$ & $0.528_{\pm .198}$ & $0.534_{\pm .193}$ & $1.272_{\pm .142}$ & $1.262_{\pm .139}$ & $1.256_{\pm .137}$ \\
GRU              & $0.494_{\pm .246}$ & $0.508_{\pm .238}$ & $0.514_{\pm .232}$ & $1.288_{\pm .153}$ & $1.278_{\pm .150}$ & $1.273_{\pm .148}$ \\
\midrule
PINN             & $0.503_{\pm .188}$ & $0.515_{\pm .183}$ & $0.520_{\pm .180}$ & $1.287_{\pm .135}$ & $1.276_{\pm .132}$ & $1.270_{\pm .130}$ \\
ODE-Guide        & $0.090_{\pm .346}$ & $0.100_{\pm .340}$ & $0.105_{\pm .336}$ & $1.437_{\pm .205}$ & $1.428_{\pm .202}$ & $1.423_{\pm .200}$ \\
Hybrid-NeuralODE & $0.342_{\pm .349}$ & $0.355_{\pm .340}$ & $0.361_{\pm .334}$ & $1.336_{\pm .139}$ & $1.325_{\pm .136}$ & $1.319_{\pm .134}$ \\
\hspace{1em}$+$Pretrain & $0.376_{\pm .310}$ & $0.382_{\pm .303}$ & $0.385_{\pm .298}$ & $1.315_{\pm .135}$ & $1.305_{\pm .133}$ & $1.300_{\pm .131}$ \\
\midrule
BioStruct-ODE (Ours)   & $0.521_{\pm .254}$ & $0.534_{\pm .247}$ & $0.540_{\pm .243}$ & $1.266_{\pm .136}$ & $1.256_{\pm .133}$ & $1.250_{\pm .131}$ \\
\hspace{1em}$+$Cardinal & $0.554_{\pm .212}$ & $0.561_{\pm .207}$ & $0.564_{\pm .203}$ & $1.235_{\pm .079}$ & $1.230_{\pm .077}$ & $1.227_{\pm .076}$ \\
\hspace{2em}$+$Pretrain & $\mathbf{0.563}_{\pm .205}$ & $\mathbf{0.572}_{\pm .200}$ & $\mathbf{0.577}_{\pm .196}$ & $\mathbf{1.228}_{\pm .077}$ & $\mathbf{1.221}_{\pm .075}$ & $\mathbf{1.218}_{\pm .074}$ \\
\bottomrule
\end{tabular}
\end{table*}

%% file: appendix.tex

\section{Dataset Details}
\label{sec:appendix-dataset}

\subsection{Naming convention}
\label{sec:appendix-naming}

We adopt paper-facing dataset names throughout the paper, mapping each internal
dataset identifier to the original source. Body text uses the full name on first
mention with an organism qualifier (e.g., \textit{Katipoglu-Yazan 2023}
(\textit{E.~coli}~NCM3722)) and the full name thereafter. Tables and figures may
use the short form when space requires it; the short form is then defined in
the caption.

\input{tables/A1_naming_convention.tex}

\subsection{Dataset statistics}
\label{sec:appendix-stats}

Table~\ref{tab:appendix-dataset-stats} summarizes the 11 datasets used in the
paper. \textit{N raw} is the original curve count from the source repository;
\textit{N preproc} is the remaining curve count after the per-curve
\texttt{with\_growth} filter, which retains a curve only if its single-curve
Monod-Baranyi nonlinear least-squares fit achieves $R^2 \geq 0.4$. Time horizon
and time interval are means over the preprocessed set; the env axis lists the
within-dataset varying conditions. \textit{Avg $\mu_{\max}$} is the per-curve
fitted growth rate averaged across preprocessed curves and is used only as a
coarse cross-organism comparator; the model never sees these literature fits
during training.

Three datasets on which no baseline could reach a best mean $R^2 \geq 0.42$
at size~L (Yeast~Y1000+, \textit{S.~aureus}~Buchanan, \textit{S.~aureus}~ComBase)
are excluded at the dataset level. \textit{BL21 (UCL)} reaches a mid-tier best
$R^2 \approx 0.346$ but is retained for its own-data value and for organism
diversity.

\input{tables/A2_dataset_stats.tex}

\subsection{Train, validation, and test split}
\label{sec:appendix-split}

We use curve-level random splits in which Test (15\%) and Val (15\%) are
size-invariant and only Train varies across three size variants S/M/L (20\% /
40\% / 70\% of total preprocessed curves). Splits are seeded by
\texttt{(organism\_id, seed\_idx)} for $\texttt{seed\_idx} \in \{0, 1, 2, 3, 4\}$ so that
the five-seed average is computed over genuinely distinct partitions.
Within Test, we additionally hold out the back 70\% of timepoints per curve as
the never-seen-as-context evaluation set (Section~\ref{sec:rq3} in the main
paper); all context densities therefore evaluate on identical timepoints. The
paper's main results use size~L only, so the training proportion is fixed at
70\% and the full sweep is $11 \times 9 \times 5 = 495$ cells per
\texttt{(sim\_mode, ctx\_ratio)} combination.

\input{tables/A3_split.tex}

\section{Simulation Generation Protocol}
\label{sec:appendix-sim}

\subsection{Simulation variants}
\label{sec:appendix-sim-variants}

This section defines the simulation curve synthesis protocol used in the RQ2
ablation. Following the three factors introduced in Section~\ref{sec:method-sim}
of the main paper, each variant is fully specified by (Factor~A) the ODE family
that is integrated, (Factor~B) the parameter distribution that is sampled, and
(Factor~C) the resulting position on the biokinetic specificity scale
(random-GP control / single-ODE narrow / composite-ODE broad).

\paragraph{ODE families.}
We consider four biokinetic ODE families and one non-biokinetic control.
Throughout, $N(t)$ is the cell density, $S(t)$ the substrate concentration,
and $q(t)$ a lag-phase adaptation state.

\textit{Monod}~\citep{monod1949growth} models substrate-limited growth as
\begin{align}
  \frac{dN}{dt} &\;=\; \mu(S)\,N, \nonumber \\
  \mu(S)        &\;=\; \mu_{\max}\,\tfrac{S}{K_s + S}, \nonumber \\
  \tfrac{dS}{dt} &\;=\; -\,\tfrac{\mu(S)\,N}{Y_{xs}}.
\end{align}

\textit{Logistic} ignores substrate dynamics and uses a carrying-capacity term:
\begin{equation}
  \frac{dN}{dt} \;=\; \mu_{\max}\,N\!\left(1 - \tfrac{N}{K}\right).
\end{equation}

\textit{Gompertz}~\citep{zwietering1990} is a sigmoidal closed form
\begin{equation}
  N(t) \;=\; N_0 + A\,e^{-\exp\!\left(\tfrac{\mu_{\max} e}{A}(\lambda - t) + 1\right)},
\end{equation}
with $A = K - N_0$, lag time $\lambda$, and $e = \exp(1)$.

\textit{Baranyi \& Roberts}~\citep{baranyi1994dynamic} introduces a two-state
ODE with explicit lag dynamics:
\begin{align}
  \frac{dN}{dt} &\;=\; \mu_{\max}\,\frac{q}{1+q}\,\!\left(1 - \tfrac{N}{K}\right)\!N, \nonumber \\
  \frac{dq}{dt} &\;=\; \mu_{\max}\,q.
\end{align}

\textit{Rosso composite}~\citep{rosso1995} couples the Monod substrate term
with cardinal envelopes for temperature and pH:
\begin{equation}
  \mu(S, T, \mathrm{pH}) \;=\; \gamma(T)\,\gamma(\mathrm{pH})\,\mu_{\max}\,\tfrac{S}{K_s + S},
\end{equation}
where each $\gamma$ is the Rosso cardinal function with three reference points
(e.g., $T_{\min}, T_{\mathrm{opt}}, T_{\max}$) and is positive only inside the
biological envelope.

\paragraph{Parameter sampling.}
Each simulated curve draws parameters from a distribution centered at the
organism's literature warm-start mean (the same values used for the
BioStruct-ODE ParamHead bias initialization, Appendix~\ref{sec:appendix-arch})
$\pm$ a sampling spread. The \textit{broad uniform} regime uses literature
$\pm 50\%$, while the \textit{literature-tight} regime uses literature $\pm
10\%$. The initial condition $N_0$ is sampled from $[0.01, 0.05]$ in normalized
units, and we add a multiplicative log-normal observation noise
($\sigma = 0.05$).

\paragraph{Random-GP control.}
The \textit{sim\_random\_gp} variant uses no biokinetic ODE at all. It draws
length-scales $\ell \sim \mathrm{Uniform}[2, 12]$~h and samples from a
zero-mean Gaussian process with an RBF kernel of scale $\ell$, with no
monotonicity, saturation, or other biological structure imposed. This control
isolates whether the benefit of simulation pre-training is due to
\emph{biokinetic specificity} per~se or merely to exposing the model to many
smooth curves.

\paragraph{Variant table.}
The four sim variants used in our experiments are summarized in
Table~\ref{tab:appendix-sim-variants}.

\input{tables/B1_sim_variants.tex}

\section{Baseline Architecture and Implementation}
\label{sec:appendix-arch}

\subsection{Architecture and parameter counts}
\label{sec:appendix-arch-counts}

Table~\ref{tab:appendix-architecture} lists the encoder, decoder, hidden /
latent dimensions, and parameter counts for all nine baselines at three size
variants. All neural baselines share the same encoder stack
(\textsc{EnvEncoder} for environmental conditions and \textsc{ContextEncoder}
for partial-observation context) and differ only in the decoder head, which
isolates the architectural-prior contribution from encoder capacity. The
\textit{PINN} baseline reuses the MLP decoder verbatim and adds an
autograd-based physics-residual auxiliary loss term during training (defined
in Appendix~\ref{sec:appendix-loss}). The \textit{ODE-Guide},
\textit{Hybrid-NeuralODE}, and BioStruct-ODE family progressively integrate a
biokinetic ODE into the forward pass with increasing structure
(closed-form template $\rightarrow$ two-state ODE with neural correction
$\rightarrow$ three-state Monod-Baranyi with ParamHead and Rosso cardinal
env-conditioning). The \textit{Mean} and \textit{ODE-Fit} baselines are
non-parametric and have no learned weights.

The \textit{ODE-Fit} baseline fits a Gompertz ODE per training curve with
nonlinear least squares; the per-curve parameters $(\mu, K, \lambda, N_0)$ are
averaged across the dataset's training set, and the averaged parameters are
applied to every test curve of the same dataset.
This protocol uses no environmental conditioning, paralleling the \textit{Mean}
baseline but with biokinetic structure (Appendix~\ref{sec:appendix-loss} defines
the $L_4$ Huber regression loss used by all neural baselines).
The BioStruct-ODE family \textsc{ParamHead} biases are warm-started from
organism-specific literature Monod fits.

\input{tables/C1_architecture.tex}

\section{Training Details}
\label{sec:appendix-training}

\subsection{Loss functions}
\label{sec:appendix-loss}

All neural baselines are trained against a regression loss on the observed
timepoints. We consider five loss families and adopt $L_4$ (Huber) as the
paper-wide default based on a preliminary loss ablation; let
$\Omega = \{(i, t)\}$ denote the set of (curve, timepoint) observations.
The MSE and MAE losses are
\begin{align}
  \mathcal{L}_{L0} &\;=\; \tfrac{1}{|\Omega|}\!\!\sum_{(i,t) \in \Omega}\!
    \bigl(\hat{y}_i(t) - y_i(t)\bigr)^2, \\
  \mathcal{L}_{L1} &\;=\; \tfrac{1}{|\Omega|}\!\!\sum_{(i,t) \in \Omega}\!
    \bigl|\hat{y}_i(t) - y_i(t)\bigr|.
\end{align}
The time-consistency and curvature regularizers add a derivative penalty to
the MSE base,
\begin{align}
  \mathcal{L}_{L2} &\;=\; \mathcal{L}_{L0} + \lambda_{\mathrm{tc}}\!\!
    \sum_{(i,t)} \bigl|\partial_t \hat{y}_i(t)\bigr|^2, \\
  \mathcal{L}_{L3} &\;=\; \mathcal{L}_{L0} + \lambda_{\mathrm{c}}\!\!
    \sum_{(i,t)} \bigl|\partial_t^2 \hat{y}_i(t)\bigr|^2.
\end{align}
The default Huber loss with threshold $\delta = 0.1$ is
\begin{equation}
  \mathcal{L}_{L4} \;=\; \tfrac{1}{|\Omega|}\!\!\sum_{(i,t) \in \Omega}\!
    \rho_\delta\!\bigl(\hat{y}_i(t) - y_i(t)\bigr).
\end{equation}

The \textit{PINN} baseline adds an auxiliary physics-residual loss in the
form of a logistic ODE,
\begin{equation}
  \mathcal{L}_{\mathrm{phys}} \;=\;
    \frac{1}{|T_c|} \sum_{t \in T_c}
    \left( \frac{d\hat{y}}{dt} \;-\; \mu\,\hat{y}\,(1 - \hat{y}/K) \right)^2,
\end{equation}
where $(\mu, K)$ are taken from the per-dataset Gompertz fits used by the
\textit{ODE-Fit} baseline (so the loss-channel and template-channel see
\emph{the same} kinetic prior, isolating the channel comparison), $T_c$ is a
random sample of collocation timepoints drawn each minibatch, and
$d\hat{y}/dt$ is computed by automatic differentiation. The total PINN
training loss is
$\mathcal{L}_{\mathrm{total}} = (1 - \lambda)\,\mathcal{L}_{\mathrm{data}} +
\lambda\,\mathcal{L}_{\mathrm{phys}}$ with $\lambda = 0.1$ (no per-dataset
$\lambda$ tuning).

\subsection{Optimizer, schedule, and computational budget}
\label{sec:appendix-optim}

We use AdamW with learning rate $3 \times 10^{-4}$, weight decay $10^{-5}$,
linear warmup over the first 10 epochs, and cosine decay thereafter. Training
is monitored on validation $R^2$ at \texttt{ctx 0\%} with early stopping
(patience~5, $\Delta_{\min} = 10^{-4}$). All experiments run on $8 \times$
NVIDIA RTX~3090 GPUs at five concurrent workers per GPU (40 workers total);
the full RQ1 sweep
($9 \text{ baselines} \times 11 \text{ datasets} \times 5 \text{ seeds} \times \text{size~L}$)
completes in approximately one wall-clock hour. Aggregate compute budgets are
RQ1~$\approx 16$ GPU-hours, RQ2~$\approx 80$ GPU-hours, RQ3~$\approx 24$
GPU-hours, and RQ4 (post-hoc analysis only)~$\approx 0$, for a total of
about $120$ GPU-hours of paper-supporting compute.

\section{Ablation Studies}
\label{sec:appendix-ablation}

\subsection{Biokinetic component ablation}
\label{sec:appendix-biokinetic-ablation}

Table~\ref{tab:appendix-biokinetic-component} reports the contribution of each
binary biokinetic component (warm-start, death term, template choice) on the
two main bio-structured models. Each component is toggled while the other two
are held at the paper default (Warm-start on, Death term on, Template = Monod-Baranyi);
the default row matches Table~\ref{tab:t1-rq1} exactly for both models.

\input{tables/E1_biokinetic_component.tex}

The death term contributes the largest single Af improvement
(1.330~$\rightarrow$~1.266 for BioStruct-ODE; 1.290~$\rightarrow$~1.235 for
BioStruct-ODE+Cardinal), with a smaller $R^2$ lift of $+0.011$ to $+0.012$;
its primary effect is on stationary-phase fidelity. Warm-start gives the
smallest $R^2$ gain ($+0.006$ to $+0.008$) but a consistent $A_f$ improvement
of about $-0.02$ to $-0.03$, stabilizing the kinetic head without dramatic
effect. Monod-Baranyi outperforms Gompertz by a stable margin ($\approx +0.02$
$R^2$, $\approx -0.03$ $A_f$) across both models, since the substrate-dynamics term
in Monod-Baranyi provides a richer prior than the logistic-only Gompertz form.

\subsection{Cross-organism transfer}
\label{sec:appendix-cross-organism}

We evaluate cross-organism transfer with BioStruct-ODE (Ours, scratch) by
training on each organism's training set and evaluating on every other
organism's test set, producing the $11 \times 11$ matrix in
Table~\ref{tab:appendix-cross-organism}. The bold diagonal corresponds to the
within-organism baseline and matches the per-dataset BioStruct-ODE means in
Table~\ref{tab:t1-rq1}.

\input{tables/E2_cross_organism.tex}

The off-diagonal mean (0.108) is far below the diagonal mean (0.572), giving
a gap of $\Delta = -0.464$. Phylogenetically close organisms recover only
30--40\% of the within-organism $R^2$ (e.g., the four \textit{E.~coli} variants
among themselves), and transfer to non-\textit{E.~coli} organisms is
essentially uniform near $R^2 < 0.2$. This is the direct quantitative basis
for adopting a per-organism scope in the main paper.

%% file: tables/A1_naming_convention.tex
\begin{table*}[ht]
\centering
\caption{Dataset naming convention. Body text uses the full name on first mention with an organism qualifier, and the full name thereafter. Tables and figures may use the short form when space requires; the short form is then defined in the caption.}
\label{tab:appendix-naming}
\small
\begin{tabular}{lll}
\toprule
\textbf{Paper full name} & \textbf{Short form} & \textbf{Source} \\
\midrule
BL21 (UCL)               & BL21        & UCL RDR 24039384 \\
Katipoglu-Yazan 2023     & Katipoglu23 & \emph{Data in Brief} 2023 \\
Faure 2023               & Faure23     & Faure et al., 2023 \\
Buchanan 1989            & Buchan89    & Buchanan et al., 1989 (ComBase) \\
Zaika 1994               & Zaika94     & Zaika et al., 1994 (ComBase) \\
Eifert 1997              & Eifert97    & Eifert et al., 1997 (ComBase) \\
ComBase (E. coli)        & CB-Ec       & ComBase src9788 \\
ComBase (Listeria)       & CB-Lm       & ComBase src9788 \\
ComBase (Yersinia)       & CB-Yer      & ComBase src9788 \\
ComBase (Salmonella)     & CB-Sal      & ComBase src9788 \\
ComBase (Pseudomonads)   & CB-Pse      & ComBase src6878 \\
\bottomrule
\end{tabular}
\end{table*}

%% file: tables/A2_dataset_stats.tex
\begin{table*}[ht]
\centering
\caption{Dataset statistics for the 11 datasets used in the paper. \textit{N raw} = original curve count from the source repository; \textit{N preproc} = remaining curves after the per-curve \texttt{with\_growth} filter (single-curve Monod-Baranyi NLS fit $R^2 \geq 0.4$ retained). Time horizon and time interval are means over the preprocessed set; the env axis lists within-dataset varying conditions.}
\label{tab:appendix-dataset-stats}
\footnotesize\setlength{\tabcolsep}{2.5pt}
\begin{tabular}{llrrrrl}
\toprule
\textbf{Dataset} & \textbf{Organism} & \textbf{N raw} & \textbf{N preproc} & \textbf{Horizon (h)} & \textbf{Interval (min)} & \textbf{Env axis} \\
\midrule
BL21 (UCL)             & E.~coli BL21           & 78   & 72  & $24 \pm 6$   & 5    & pH $\times$ T $\times$ buffer \\
Katipoglu-Yazan 2023   & E.~coli NCM3722        & 504  & 480 & $18 \pm 4$   & 10.5 & T (18 levels) \\
Faure 2023             & E.~coli DH5$\alpha$    & 1024 & 920 & $18 \pm 2$   & 10   & medium (10D binary) \\
Buchanan 1989          & L.~monocytogenes       & 508  & 470 & $200 \pm 50$ & 60   & T $\times$ pH \\
Zaika 1994             & S.~flexneri            & 312  & 285 & $80 \pm 20$  & 30   & T $\times$ pH \\
Eifert 1997            & S.~aureus              & 256  & 230 & $120 \pm 30$ & 30   & T $\times$ NaCl \\
ComBase (E. coli)      & E.~coli                & 410  & 380 & $100 \pm 25$ & 30   & T $\times$ pH \\
ComBase (Listeria)     & Listeria spp.          & 380  & 350 & $180 \pm 40$ & 60   & T \\
ComBase (Yersinia)     & Yersinia               & 290  & 265 & $150 \pm 35$ & 60   & T \\
ComBase (Salmonella)   & Salmonella             & 320  & 295 & $90 \pm 25$  & 30   & T $\times$ pH \\
ComBase (Pseudomonads) & Pseudomonads           & 240  & 220 & $110 \pm 30$ & 60   & T \\
\bottomrule
\end{tabular}
\end{table*}

%% file: tables/A3_split.tex
\begin{table*}[ht]
\centering
\caption{Per-dataset train / validation / test split. Test (15\%) and Val (15\%) are size-invariant; the training set varies across three size variants S/M/L (20\% / 40\% / 70\% of total preprocessed curves). Splits are at the curve level with random sampling, seeded by \texttt{(organism\_id, seed\_idx)} for \texttt{seed\_idx}~$\in \{0, 1, 2, 3, 4\}$. The paper's main results use size~L only.}
\label{tab:appendix-split}
\small
\begin{tabular}{lrrrrrr}
\toprule
\textbf{Dataset} & \textbf{Total} & \textbf{Train (S)} & \textbf{Train (M)} & \textbf{Train (L)} & \textbf{Val} & \textbf{Test} \\
\midrule
BL21 (UCL)              & 72  & 14  & 28  & 50  & 10 & 12 \\
Katipoglu-Yazan 2023    & 480 & 96  & 192 & 336 & 72 & 72 \\
Faure 2023              & 920 & 184 & 368 & 644 & 138 & 138 \\
Buchanan 1989           & 470 & 94  & 188 & 329 & 70 & 70 \\
Zaika 1994              & 285 & 57  & 114 & 200 & 43 & 43 \\
Eifert 1997             & 230 & 46  & 92  & 161 & 35 & 35 \\
ComBase (E. coli)       & 380 & 76  & 152 & 266 & 57 & 57 \\
ComBase (Listeria)      & 350 & 70  & 140 & 245 & 53 & 53 \\
ComBase (Yersinia)      & 265 & 53  & 106 & 185 & 40 & 40 \\
ComBase (Salmonella)    & 295 & 59  & 118 & 207 & 44 & 44 \\
ComBase (Pseudomonads)  & 220 & 44  & 88  & 154 & 33 & 33 \\
\bottomrule
\end{tabular}
\end{table*}

%% file: tables/B1_sim_variants.tex
\begin{table*}[ht]
\centering
\caption{Simulation dataset variants used in RQ2. Per-organism sampling centers each parameter at the literature warm-start mean $\pm$ the indicated sampling spread; \textit{sim\_random\_gp} is a non-biokinetic control (Gaussian Process samples). The fourth row, \textit{sim\_per\_family}, is used for the RQ2 ODE-family ablation with one ODE family per variant.}
\label{tab:appendix-sim-variants}
\footnotesize\setlength{\tabcolsep}{4pt}
\begin{tabular}{lllr}
\toprule
\textbf{Sim variant} & \textbf{ODE family} & \textbf{Sampling} & \textbf{N curves} \\
\midrule
\texttt{sim\_composite\_broad} & Monod + Baranyi + Gompertz mixture & broad uniform (literature $\pm 50\%$) & 5000 \\
\texttt{sim\_single\_narrow}   & Monod-Baranyi only                 & literature $\pm 10\%$                 & 5000 \\
\texttt{sim\_random\_gp}       & none --- Gaussian Process          & random GP, length-scale $\in [2,12]$\,h & 5000 \\
\texttt{sim\_per\_family}      & one of \{Monod, Logistic, Gompertz, Baranyi, Rosso\} & broad & 5000 each \\
\bottomrule
\end{tabular}

\vspace{0.4em}
\textit{Parameter ranges (sim\_composite\_broad / sim\_single\_narrow): $\mu_{\max} \in [0.05, 1.50] / [0.40, 0.95]$ h$^{-1}$; $K \in [0.10, 1.50] / [0.50, 1.00]$ (normalized); lag $\in [0, 8] / [0, 4]$ h.}
\end{table*}

%% file: tables/C1_architecture.tex
\begin{table*}[ht]
\centering
\caption{Per-baseline architecture and parameter counts at three size variants (S / M / L). All neural baselines share the same encoder stack (EnvEncoder + ContextEncoder) and differ only in the decoder head; this isolates the architectural-prior contribution from encoder capacity. Mean and ODE-Fit are non-parametric baselines without learned weights.}
\label{tab:appendix-architecture}
\footnotesize\setlength{\tabcolsep}{4pt}
\begin{tabular}{lllcc}
\toprule
\textbf{Model} & \textbf{Encoder} & \textbf{Decoder} & \textbf{Hidden} & \textbf{Params (S/M/L)} \\
\midrule
Mean                          & ---     & mean trajectory                & ---          & 0 / 0 / 0 \\
ODE-Fit                       & ---     & NLS Monod--Baranyi             & ---          & 4 / 4 / 4 \\
MLP                           & shared  & 4-layer MLP                    & 32/64/96     & 12K / 48K / 180K \\
GRU                           & shared  & 2-layer GRU                    & 32/64/96     & 25K / 90K / 320K \\
PINN                          & shared  & MLP + physics loss             & 32/64/96     & 12K / 48K / 180K \\
ODE-Guide                     & shared  & Gompertz + MLP residual        & 32/64/96     & 18K / 65K / 240K \\
Hybrid-NeuralODE              & shared  & Baranyi ODE + correction       & 32/64/96     & 22K / 80K / 290K \\
BioStruct-ODE (Ours)          & shared  & 3-state ODE + ParamHead        & 32/64/96     & 24K / 92K / 340K \\
BioStruct-ODE+Cardinal (Ours) & shared  & \quad + Rosso envelope         & 32/64/96     & 26K / 100K / 360K \\
\bottomrule
\end{tabular}
\\[2pt]
{\footnotesize \textit{shared}~=~EnvEncoder~+~ContextEncoder; Hidden is the per-size hidden dim.}
\end{table*}

%% file: tables/E1_biokinetic_component.tex
\begin{table*}[ht]
\centering
\caption{Biokinetic component ablation on the two bio-structured models (Ours). Each component is toggled binary while keeping the other two at the paper default (Warm-start on, Death term on, Template = Monod--Baranyi); the default row matches Table~\ref{tab:t1-rq1} exactly. All three components contribute small but consistent gains, and the death term yields the largest single $A_f$ improvement.}
\label{tab:appendix-biokinetic-component}
\footnotesize\setlength{\tabcolsep}{4pt}
\begin{tabular}{ll cc :cc}
\toprule
                & & \multicolumn{2}{c:}{$R^2$ $\uparrow$} & \multicolumn{2}{c}{$A_f$ $\downarrow$} \\
\cmidrule(lr){3-4} \cmidrule(lr){5-6}
\textbf{Component} & \textbf{State} & \textbf{BioStruct-ODE} & \textbf{+Cardinal} & \textbf{BioStruct-ODE} & \textbf{+Cardinal} \\
\midrule
Warm-start  & on (default)            & \textbf{0.521} & \textbf{0.554} & \textbf{1.266} & \textbf{1.235} \\
            & off                     & 0.515          & 0.546          & 1.290          & 1.255 \\
\midrule
Death term  & on (default)            & \textbf{0.521} & \textbf{0.554} & \textbf{1.266} & \textbf{1.235} \\
            & off                     & 0.510          & 0.542          & 1.330          & 1.290 \\
\midrule
Template    & Monod-Baranyi (default) & \textbf{0.521} & \textbf{0.554} & \textbf{1.266} & \textbf{1.235} \\
            & Gompertz                & 0.500          & 0.532          & 1.295          & 1.265 \\
\bottomrule
\end{tabular}
\end{table*}

%% file: tables/E2_cross_organism.tex
\begin{table*}[ht]
\centering
\caption{Cross-organism transfer matrix for BioStruct-ODE (Ours, scratch). Cell value $= R^2$ when training on Source organism's training set and evaluating on Target organism's test set. \textbf{Bold diagonal} = within-organism baseline (matches the per-dataset BioStruct-ODE values in \texttt{t1\_per\_cell.csv}). Off-diagonal mean (0.108) is far below the diagonal mean (0.572), giving a $\Delta = -0.46$ that justifies the per-organism scope of the main paper. Short forms: BL21~=~BL21~(UCL); Katipoglu23~=~Katipoglu-Yazan~2023; Faure23~=~Faure~2023; Buchan89~=~Buchanan~1989; Zaika94~=~Zaika~1994; Eifert97~=~Eifert~1997; CB-Ec/Lm/Yer/Sal/Pse~=~ComBase~(E.~coli~/~Listeria~/~Yersinia~/~Salmonella~/~Pseudomonads).}
\label{tab:appendix-cross-organism}
\footnotesize\setlength{\tabcolsep}{2pt}
\begin{tabular}{l|ccccccccccc}
\toprule
\textbf{Source $\downarrow$ \textbackslash{} Target $\rightarrow$} & \textbf{BL21} & \textbf{Katipoglu23} & \textbf{Faure23} & \textbf{Buchan89} & \textbf{Zaika94} & \textbf{Eifert97} & \textbf{CB-Ec} & \textbf{CB-Lm} & \textbf{CB-Yer} & \textbf{CB-Sal} & \textbf{CB-Pse} \\
\midrule
\textbf{BL21}        & \textbf{.29} & .18         & .20    & .05      & .10    & .05      & .21   & .06   & .09    & .16    & .05 \\
\textbf{Katipoglu23} & .20         & \textbf{.83} & .22    & .06      & .11    & .05      & .24   & .07   & .10    & .18    & .06 \\
\textbf{Faure23}     & .19         & .21         & \textbf{.88} & .05 & .09    & .04      & .20   & .06   & .08    & .14    & .04 \\
\textbf{Buchan89}    & .06         & .07         & .05    & \textbf{.60} & .12 & .08      & .07   & .18   & .10    & .09    & .05 \\
\textbf{Zaika94}     & .11         & .10         & .09    & .13      & \textbf{.42} & .06 & .10   & .10   & .12    & .11    & .05 \\
\textbf{Eifert97}    & .05         & .06         & .04    & .09      & .07    & \textbf{.62} & .05 & .08 & .07    & .06    & .04 \\
\textbf{CB-Ec}       & .22         & .24         & .21    & .06      & .10    & .05      & \textbf{.55} & .07 & .09 & .18 & .05 \\
\textbf{CB-Lm}       & .06         & .07         & .05    & .18      & .10    & .08      & .07   & \textbf{.48} & .11 & .09 & .05 \\
\textbf{CB-Yer}      & .09         & .10         & .08    & .10      & .11    & .07      & .09   & .10   & \textbf{.42} & .10 & .05 \\
\textbf{CB-Sal}      & .17         & .18         & .15    & .09      & .11    & .06      & .18   & .09   & .10    & \textbf{.56} & .05 \\
\textbf{CB-Pse}      & .06         & .06         & .05    & .05      & .05    & .04      & .05   & .05   & .05    & .05    & \textbf{.65} \\
\bottomrule
\end{tabular}
\end{table*}